\documentclass[sigconf]{acmart}
\AtBeginDocument{%
  }
    
\usepackage[english]{babel} % 选择适当的语言      
\usepackage{graphicx}
\usepackage{amsmath}
\usepackage{booktabs}

\usepackage{multirow}
\usepackage[skip=3pt]{caption} % 设置 caption 与正文的间距
\usepackage{makecell}
\setcopyright{none}
\newif\ifincludeappendix
\includeappendixfalse  % 如果不想包含附录就设为 false

\makeatletter
\ifdefined\includeappendix
  \ifincludeappendix
  \else
    \@ifundefined{r@apen.visual}{\newlabel{apen.visual}{{A}{0}}}{}
    \@ifundefined{r@apen.weights}{\newlabel{apen.weights}{{B}{0}}}{}
    \@ifundefined{r@apen.a}{\newlabel{apen.a}{{C}{0}}}{}
    \@ifundefined{r@apen.b}{\newlabel{apen.b}{{D}{0}}}{}
  \fi
\else
  \@ifundefined{r@apen.visual}{\newlabel{apen.visual}{{A}{0}}}{}
  \@ifundefined{r@apen.weights}{\newlabel{apen.weights}{{B}{0}}}{}
  \@ifundefined{r@apen.a}{\newlabel{apen.a}{{C}{0}}}{}
  \@ifundefined{r@apen.b}{\newlabel{apen.b}{{D}{0}}}{}
\fi
\makeatother
\copyrightyear{2025}
\acmYear{2025}
\setcopyright{cc}
\setcctype{by}
\acmConference[MMAsia '25]{ACM Multimedia Asia}{December 9--12, 2025}{Kuala Lumpur, Malaysia}
\acmBooktitle{ACM Multimedia Asia (MMAsia '25), December 9--12, 2025, Kuala Lumpur, Malaysia}\acmDOI{10.1145/3743093.3771019}
\acmISBN{979-8-4007-2005-5/2025/12}

\author{Shiyi Zhang}
\orcid{0009-0008-6165-5496}
\affiliation{%
  \institution{Department of Electronic and Electrical Engineering, Southern University of Science and Technology}
  \city{Shenzhen}
  \country{China}
}
\email{sy.zhang4@siat.ac.cn}

\author{Dong Liang}
\affiliation{%
  \institution{Shenzhen Institutes of Advanced Technology, Chinese Academy of Sciences}
  \city{Shenzhen}
  \country{China}
}
\email{dong.liang@siat.ac.cn}

\author{Yihang Zhou}
\authornote{Corresponding Author}
\affiliation{%
  \institution{Shenzhen Institutes of Advanced Technology, Chinese Academy of Sciences}
  \city{Shenzhen}
  \country{China}
}
\email{yh.zhou2@siat.ac.cn}

\renewcommand{\shortauthors}{Shiyi et al.}

\begin{document}

% 定义HAVIR字母的配色（可自定义）

\title{NeuroSwift: A Lightweight Cross-Subject Framework
for fMRI Visual Reconstruction of Complex Scenes}



\begin{abstract}
Reconstructing visual information from brain activity via computer vision technology provides an intuitive understanding of visual neural mechanisms. Despite progress in decoding fMRI data with generative models, achieving accurate cross‑subject reconstruction of visual stimuli remains challenging and computationally demanding. This difficulty arises from inter‑subject variability in neural representations and the brain’s abstract encoding of core semantic features in complex visual inputs.

To address these challenges, we propose NeuroSwift, which integrates complementary adapters via diffusion: AutoKL for low-level features and CLIP for semantics. NeuroSwift's CLIP Adapter is trained on Stable Diffusion generated images paired with COCO captions to emulate higher visual cortex encoding. For cross-subject generalization, we pretrain on one subject then fine-tune only 17\% parameters (FC layers) for new subjects, freezing other components. This enables state-of-the-art performance with only 1-hour training per subject on lightweight GPUs (three RTX4090), outperforming existing methods.
\end{abstract}




\maketitle

\section{Introduction}
\label{sec:intro}
Human brain processes complex visual information efficiently, establishing multi-level neural representations \cite{brain1,brain2,brain3,retina}. Early approaches used VAEs or GANs for image synthesis \cite{2-7,3-8,4-9,5-10,6-11,7-12} but lacked semantic accuracy. Following the NSD dataset \cite{8-14}, Takagi et al. \cite{9-1} pioneered high-resolution reconstructions using diffusion models. Subsequent works \cite{10-2,11-3-23,12-4} and recent models \cite{mindeye2,mindtuner,BrainGuard,Neuropictor} employed conditional diffusion and cross-subject decoding to enhance semantic integration and generalizability.
\begin{figure}
  \centering
  \includegraphics[width=0.45\textwidth]{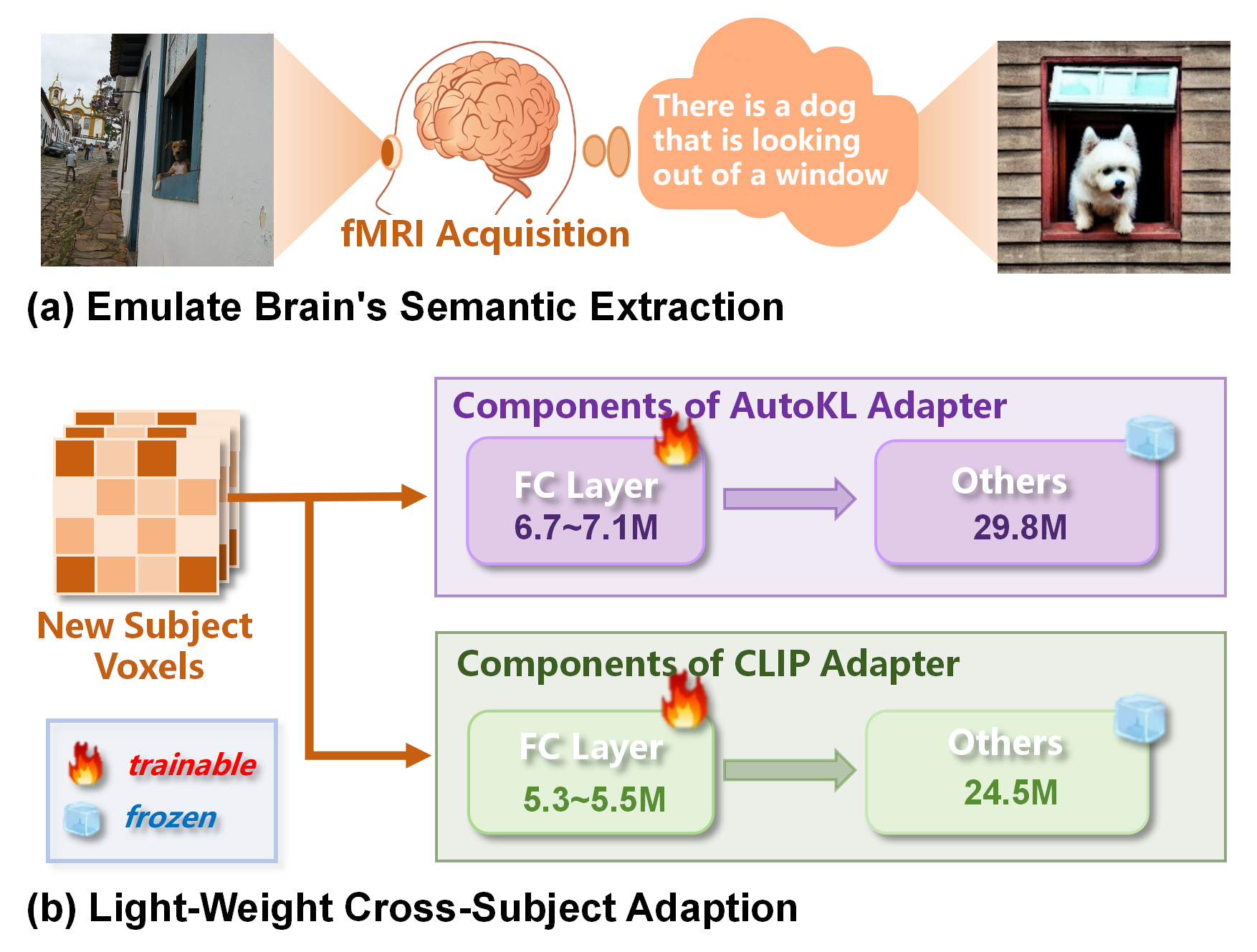}
  \caption{\textbf{(a)} During scanning, subjects imagine semantic content rather than raw visual stimuli. Therefore, we leverage COCO Captions to emulate imagined semantics and Semantic Images to emulate imagined scenes. \textbf{(b)} Efficient cross‑subject adaptation. Pretrain on the first subject, then fine‑tune only 17\% of parameters on the other subjects in one hour with 3×RTX4090 GPUs.}
  \label{fig:1}
\end{figure}

However, existing methods generally suffer from two limitations. First, although models such as MindEye2\cite{mindeye2} (64.4M parameters) and MindTuner\cite{mindtuner} (76.6M parameters) have achieved cross-subject decoding capabilities, they demand high computational resources. For example, training MindEye2 requires eight NVIDIA A100 GPUs. Nevertheless, MindEye2’s reconstruction performance in cross‑subject generalization still falls short of the results it achieves when trained individually on each subject. Second, in complex visual scenes involving cluttered backgrounds, occlusions, small objects, and dense spatial structures, existing methods often struggle to reconstruct low-level details accurately while also capturing high-level semantic information.

To mitigate limitations in complex visual scenes, our framework employs COCO captions to guide semantic extraction: text embeddings are generated by a frozen Text Clipper, while image embeddings are obtained by synthesizing semantic images via Stable Diffusion followed by feature extraction with a frozen Image Clipper. For cross-subject generalization, we pretrain on Subj01 then fine-tune only the fully connected layers (17\% of parameters) in adapters for other subjects (Subj02/05/07), freezing other components. This reduces computational costs, requiring just three NVIDIA RTX4090 GPUs. With only one hour of training data, our method surpasses existing approaches, achieving efficient state-of-the-art cross-subject visual decoding under limited resources. Our contributions are summarized as follows:
\begin{itemize}
\item We propose a biologically inspired CLIP Adapter training mechanism that jointly processes synthetically generated semantic images and textual captions to emulate the brain's efficient semantic extraction. 
    
\item We introduce a cross-subject adaptation strategy where only 17\% of lightweight FC layers are fine-tuned for new subjects, freezing all other components. This approach achieves state-of-the-art generalization capability with minimal resources, reducing computational costs compared to existing methods.
    
\item Considering neuroanatomical variability, we employ individualized brain region masks with manually delineated ROI boundaries rather than the standardized templates typically used in previous research. This enables precise brain decoding for specific subjects.
\end{itemize}

\section{Related Work}
\label{sec:relatedwork}
\subsection{Diffusion Models}
Diffusion Models (DMs) \cite{3-7,mdb8,Mdb13,14-6,4-9,11-3-23} generate samples through iterative denoising. Latent Diffusion Models (LDMs) operate in compressed latent spaces to reduce computational costs, with Stable Diffusion (SD) \cite{mdb30} being a widely-adopted LDM that leverages large-scale image-text datasets \cite{3-28} to achieve superior text-to-image performance \cite{mdb26,5-10,mdb47}. We employ SD to transform captions into semantic images. For final reconstruction, we use Versatile Diffusion (VD) \cite{mdb46} which employs a dual-guidance mechanism fusing CLIP features during denoising, trained on Laion2B-en \cite{uni31} and COYO-700M \cite{uni4}, selected for its ability to integrate multimodal embeddings.

\subsection{fMRI Visual Reconstruction}
Decoding visual stimuli from brain activity progressed from pre-deep learning methods using image features like multi-scale local image bases\cite{11-3-08,add3} and Gabor filters\cite{11-2} with linear mappings, though limited by fMRI's low signal-to-noise ratio and small datasets. Deep neural networks later modeled nonlinear relationships \cite{11-8} through architectures including deep belief networks\cite{10-4}, VAEs\cite{10-5-effenet}, feedforward networks\cite{10-6,10-7}, GANs\cite{10-8,4-9}, and hybrid VAE/GANs\cite{7-12}. While these improved pixel-level reconstruction, they lacked semantic content until CLIP integration: Lin et al.\cite{2-7} aligned fMRI patterns with CLIP's latent space via contrastive-adversarial learning, subsequently using StyleGAN2 to enhance semantic accuracy.
\begin{figure*}[h]
  \centering
  \includegraphics[width=0.7 \linewidth]{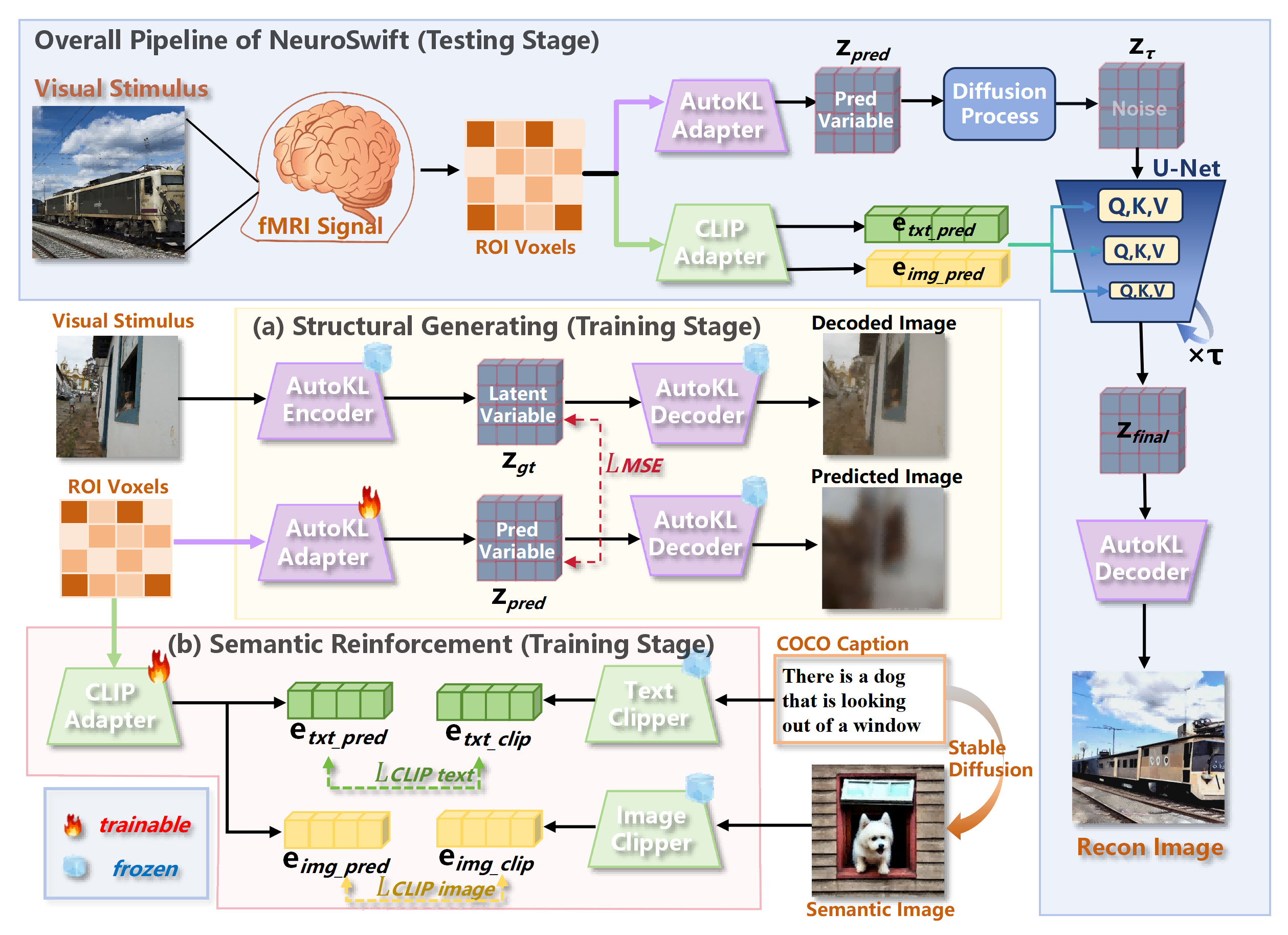}
  \caption{Overall structure of NeuroSwift in single‑subject mode. fMRI voxels are processed through hierarchical pipelines: \textbf{(a)} The structural generation pipeline transforms voxels into latent space representations($\mathit{z}_{\scriptstyle{\textit{pred}}}$), which serve as the diffusion prior for noise addition($\mathit{z}_{\scriptstyle{\tau}}$). \textbf{(b)} The semantic reinforcement pipeline projects voxels into CLIP's text ($\mathit{e}_{\scriptstyle{\textit{txt\_pred}}}$) and image ($\mathit{e}_{\scriptstyle{\textit{img\_pred}}}$) embeddings. Then, the denoising UNet iteratively refines the noise representation $\mathit{z}_{\scriptstyle{\tau}}$ to integrate the CLIP semantic embeddings.  Finally, the frozen AutoKL Decoder decodes $\mathit{z}_{\scriptstyle{\textit{final}}}$ into the reconstructed image.}
  \label{fig:2}
\end{figure*}

Since Stable Diffusion's successful application in various generative tasks\cite{3-27,3-28,3-29,3-7,add2,add3}, Takagi et al.\cite{9-1} pioneered mapping fMRI signals to diffusion latent space and CLIP text embeddings to generate images, though reconstructions lacked sufficient semantic information and natural qualities. Later, MindEye\cite{12-4}, which optimized the semantic representations of fMRI features through contrastive learning, and MindDiffuser\cite{14-6}, which devised a two-stage diffusion process, addressed this issue. 

Recent advances in cross-subject decoding include MindEye2\cite{mindeye2} enabling cross-subject alignment, MindTuner\cite{mindtuner} bridging semantic gaps via visual fingerprinting and fMRI-to-text alignment, and BrainGuard\cite{BrainGuard} supporting privacy-preserving collaborative decoding. However, these methods still struggle to balance low-level structural feature extraction with high-level semantic understanding, particularly in complex visual scenes under cross-subject generalization, while requiring substantial computational resources.

\section{NeuroSwift}
In this section, we propose a hierarchical model termed NeuroSwift for vision reconstruction, and its framework is shown in Figure~\ref{fig:2}. The \textbf{structural generation} pipeline converts voxels into the latent space of Versatile Diffusion, which function as the diffusion prior. The \textbf{semantic reinforcement} pipeline aligns voxels with CLIP's text and image embeddings, which serve as conditional guidance for the diffusion process. Finally, the \textbf{diffusion reconstruction} process synthesizes the final image by integrating the structural prior and CLIP embeddings.

\subsection{Structural Generation}
The AutoKL Adapter transforms fMRI voxels to the latent space $\mathit{z}_{\scriptstyle{\textit{pred}}}$, and the frozen pre-trained AutoKL Decoder decodes it into a predicted image. We call it "structural generation" because it captures fundamental spatial structures and color distributions that serve as a latent diffusion prior, but it lacks detailed semantic content. See Appendix~\ref{apen.a} for details of the AutoKL Adapter components.

The AutoKL Adapter is trained by minimizing the mean squared error between the predicted latent variable  \(\mathit{z}_{\scriptstyle{\textit{pred}}}\) and the ground-truth latent variable \(\mathit{z}_{\scriptstyle{\textit{gt}}}\), which is encoded from the visual stimulus using the frozen AutoKL Encoder:
\begin{equation}  
    \mathcal{L}_{\textit{MSE}} = \frac{1}{n} \sum_{i=1}^{n} \left( \mathit{z}^{\scriptstyle{(i)}}_{\scriptstyle{\textit{pred}}} - \mathit{z}^{\scriptstyle{(i)}}_{\scriptstyle{\textit{gt}}} \right)^2
\end{equation}

where \(n\) is the batch size.

\subsection{Semantic Reinforcement}
We designed a CLIP Adapter to map fMRI voxels into CLIP image and text embeddings, implementing semantic reinforcement that captures core semantics rather than structural features to condition the denoising process. The CLIP Adapter aligns predicted text embeddings $\mathit{e}_{\scriptstyle{\textit{txt\_pred}}}$ with $\mathit{e}_{\scriptstyle{\textit{txt\_clip}}}$, where the latter is decoded by a pre-trained Text Clipper from textual captions corresponding to visual stimuli. Simultaneously, it aligns predicted image embeddings $\mathit{e}_{\scriptstyle{\textit{img\_pred}}}$ with $\mathit{e}_{\scriptstyle{\textit{img\_clip}}}$, which is decoded by the pre-trained Image Clipper from semantic images generated by Stable Diffusion using the corresponding textual captions.

We use semantic images instead of original visual stimuli for CLIP image alignment because complex natural scenes often contain irrelevant content that may interfere with semantic extraction. In contrast, semantic images generated from textual captions retain only core information, thus reinforcing effectiveness. See Appendix~\ref{apen.b} for details of the CLIP Adapter components.
 
\textbf{Training objectives.}
 The CLIP Adapter framework employs two types of loss functions to optimize the mapping of CLIP image and text embeddings. Among them, the \textit{SoftCLIP loss}\cite{12-4,13-5} has demonstrated effectiveness in aligning the fMRI modality with the pretrained CLIP embedding space. Within this framework, the loss function leverages a contrastive learning mechanism to maximize the similarity of positive pairs while minimizing the similarity of negative pairs.
\begin{equation}
\small
\begin{aligned}
\mathcal{L}_{SoftCLIP}(p, t) = - \sum_{i=1}^N \sum_{j=1}^N  
\Bigg[ &\frac{\exp(t_i \cdot t_j / \tau)}
{\sum_{m=1}^N \exp\left( \frac{t_i \cdot t_m}{\tau} \right)} &\\\times\log\left( \frac{\exp(p_i \cdot t_j / \tau)}
{\sum_{m=1}^N \exp\left( \frac{p_i \cdot t_m}{\tau} \right)} \right) \Bigg]
\end{aligned}
\end{equation}

Where \textit{p}, \textit{t} are the predicted CLIP embedding and target CLIP embedding in a batch of size \textit{N}, respectively. $\tau$ is a temperature hyperparameter. However, using only the SoftCLIP loss causes noticeable artifacts because batch-wise soft labels create random variations, particularly with sparse and noisy fMRI-CLIP mappings. To address this, we introduced an MSE regularization term to ensure direct consistency between predicted and target CLIP embeddings. The complete set of losses for predicting image and text CLIP embeddings incorporating these two losses:
\begin{equation}
\begin{split}
\mathcal{L}_{CLIP image} = \mathcal{L}_{SoftCLIP}( \mathit{e}_{\scriptstyle{\textit{img\_pred}}} , \mathit{e}_{\scriptstyle{\textit{img\_clip}}} )\\+ \mathcal{L}_{MSE}( \mathit{e}_{\scriptstyle{\textit{img\_pred}}} , \mathit{e}_{\scriptstyle{\textit{img\_clip}}} )
\end{split}
\end{equation}
\begin{equation}
\begin{split}
\mathcal{L}_{CLIP text} = \mathcal{L}_{SoftCLIP}( \mathit{e}_{\scriptstyle{\textit{txt\_pred}}} , \mathit{e}_{\scriptstyle{\textit{txt\_clip}}} )\\ + \mathcal{L}_{MSE}( \mathit{e}_{\scriptstyle{\textit{txt\_pred}}} , \mathit{e}_{\scriptstyle{\textit{txt\_clip}}} )
\end{split}
\end{equation}

Here, $\mathit{e}_{\scriptstyle{\textit{txt\_clip}}}$ and $\mathit{e}_{\scriptstyle{\textit{img\_clip}}}$ represent the CLIP embeddings of text captions and their corresponding semantic images, where the captions are obtained using the COCO id of the visual stimuli, the semantic images are generated by Stable Diffusion using the captions.

\subsection{Diffusion Reconstruction}
The diffusion process synthesizes images that preserve structural features while maintaining semantic information. 

\textbf{Noise controlled initialization.} The process initializes from the latent variable $\mathit{z}_{\scriptstyle{\textit{pred}}}$ and adds partial noise controlled by the structural strength coefficient $\mathit{s} \in (0,1]$. The initial noise step $\tau = N - \lfloor N \cdot s \rfloor \quad$($N$ is the total timesteps) determines the noise level for denoising. The noised latent $\mathit{z}_{\tau}$ is derived in one step:  
\begin{equation}
\mathit{z}_{\tau} = \sqrt{\alpha_{\tau}} \mathit{z}_{\scriptstyle{\textit{pred}}} + \sqrt{1 - \alpha_{\tau}} \epsilon, \quad \epsilon \sim \mathcal{N}(0, \mathit{I})
\end{equation}
where $\alpha_{\tau}$ is defined by the noise scheduler, $\epsilon$ is stochastic noise sampled from a standard normal distribution. 

\textbf{Semantic conditioning.} The semantic constraints are imposed via cross-attention mechanisms, which guides the UNet during denoising. At each timestep $\tau_t$, the UNet predicts noise:  
\begin{equation}
\resizebox{0.4\textwidth}{!}{$
\hat{\epsilon}_{\theta}(\mathit{z}_t, \tau_t, \mathit{e}_{\textit{input}}) = 
\operatorname{UNet} \left( \mathit{z}_t, \tau_t, 
\operatorname{CrosAtt}(\mathit{e}_{\textit{txt\_pred}}, \mathit{e}_{\textit{img\_pred}} )\right)$}
\end{equation}

\textbf{Guided denoising.}  
The denoising process iteratively refines the latent vector $\mathit{z}_{\tau}$ through a sequence of timesteps $ t \in [\tau, 0] $. At each step $t$, the predicted noise $\hat{\epsilon}_{\theta}$ is used to update the latent state:
\begin{equation}
\begin{aligned}
\mathbf{z}_{t-1} = 
\frac{1}{\sqrt{\alpha_t}} \left( 
\mathbf{z}_t - \frac{\beta_t}{\sqrt{1-\bar{\alpha}_t}} \hat{\epsilon}_\theta(\mathbf{z}_t, t, \mathbf{e}_{\textit{input}})
\right) \\ + \sqrt{\beta_t} \cdot \boldsymbol{\epsilon}
\end{aligned}
\end{equation}
where $\beta_{t}$ is defined by the noise scheduler, $\alpha_t = 1 - \beta_t$, $\bar{\alpha}_t$ is the cumulative product of signal retention coefficients. This update rule progressively removes noise while integrating semantic information guided by the multimodal condition $\mathit{e}_{\scriptstyle{\textit{input}}}$.

\textbf{Image reconstruction.}
Finally, $\mathit{z}_{\scriptstyle{\textit{final}}}$ is decoded into pixel space via the AutoKL Decoder $\mathcal{D}$, yielding the reconstructed image:  
\begin{equation}
\mathit{I}_{\scriptstyle{\textit{recon}}} = \mathcal{D}( \mathit{z}_{\scriptstyle{\textit{final}}} )
\end{equation}
This diffusion process achieves structure-semantics decoupling: $\mathit{z}_{\scriptstyle{\textit{pred}}}$ provides blurred low-level visual primitives, while $\mathit{e}_{\scriptstyle{\textit{input}}}$ enforces high-level semantic concepts.

\section{Results and Analyses}
\subsection{Data Elaboration}
We utilize the Natural Scenes Dataset (NSD) \cite{8-14}, a 7T fMRI dataset of eight subjects viewing COCO images \cite{imagenet-alex}. Following established methodologies \cite{3-8,11-3-23,10-2,12-4,9-1,mdb47}, we analyze four subjects (Subj01, 02, 05, 07) with complete sessions. Our design includes: (a) 982-image shared test set; (b) subject-specific training/validation sets (8,859 images each). Preprocessed fMRI (1.8mm resolution) was masked using NSD-provided ROIs to extract ventral visual cortex voxels. The details of the data used in our experiments have been summarized in Appendix~\ref{apen.visual}.

\textbf{Notably}, previous studies have primarily relied on standardized brain templates for spatial parcellation. However, both neuroanatomical architecture and functional organization demonstrate significant inter-individual variability\cite{visualdiff}. During spatial normalization, these inter-subject discrepancies can induce boundary blurring or registration inaccuracies, ultimately compromising model training efficacy and image reconstruction performance. In contrast to conventional approaches, our study utilizes NSD provided brain masks incorporating manually delineated ROI boundaries for individual subjects. These customized masks account for neuroanatomical and functional specificity through fine grained regional mapping, enabling differentiation of areas with subtle functional distinctions, thereby enhancing downstream processing precision.

\subsection{Image Reconstruction Examples}
To assess the effectiveness of our method in reconstructing semantic details and spatial structures in complex visual scenes, we followed the NeuroSwift implementation pipeline (Figure~\ref{fig:2}) and selected four stimuli with high semantic complexity for evaluation. 
\begin{figure}[h]
  \centering
  \includegraphics[width=0.48\textwidth]{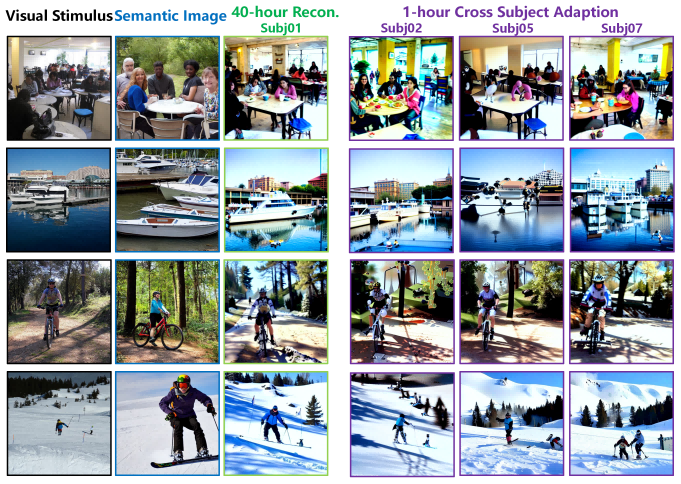}
  \caption{Examples of NeuroSwift reconstructions from complex visual stimuli.}
  \label{fig:3}
\end{figure}

As shown in Figure~\ref{fig:3}, the figure consists of six columns: the first displays the original visual stimuli (ground truth), the second shows semantically refined images generated from COCO captions, the third presents reconstructions based on 40 hours of training data from Subj01, and the right three columns illustrate cross-subject generalization results, each obtained using only one hour of training data from three different subjects.

NeuroSwift effectively reconstructs the spatial layout and semantics of complex visual scenes. For example, in the final stimulus showing a small skier gliding through snowy pine woods, it accurately recovers key scene details via a hierarchical pipeline of spatial feature extraction and semantic enhancement.

Under cross-subject generalization, NeuroSwift also reconstructs complex images with high fidelity, including scenes such as “people dining,” “two boats,” and “a person cycling in a forest.” Notably, these results are achieved with training on just 3×RTX4090 GPUs, highlighting its superior efficiency and generalization capability over existing methods.

\subsection{Qualitative Comparison}
In the qualitative experiments, we first compare the reconstruction results of NeuroSwift, fine-tuned with 1 hour of training data under the cross-subject generalization setting, against MindTuner\cite{mindtuner} and MindEye2\cite{mindeye2} under the same condition. We then compare our single-subject reconstruction results with those of MindBridge\cite{13-5}, trained with 40 hours of data under the single-subject setting.
\begin{figure}[h]
  \centering
  \includegraphics[width=0.49\textwidth]{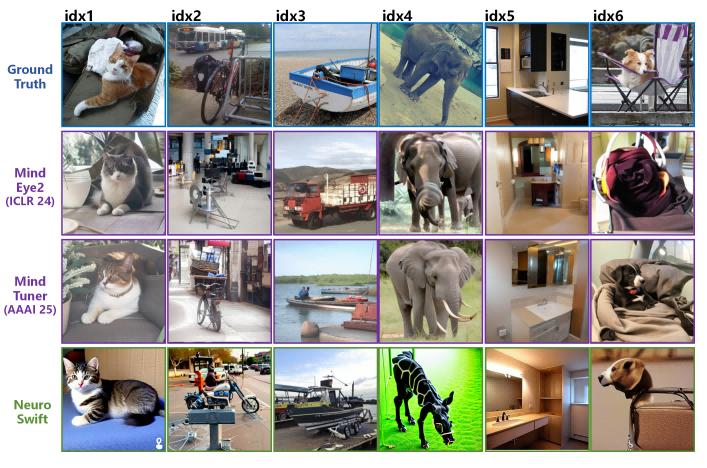}
  \caption{Comparison of our framework, MindEye2\cite{mindeye2}, and MindTuner\cite{mindtuner} using 1h of training data in \textbf{cross‑subject} adaptation mode.}
  \label{fig:4}
\end{figure}
\begin{figure}
  \centering
  \includegraphics[width=0.49\textwidth]{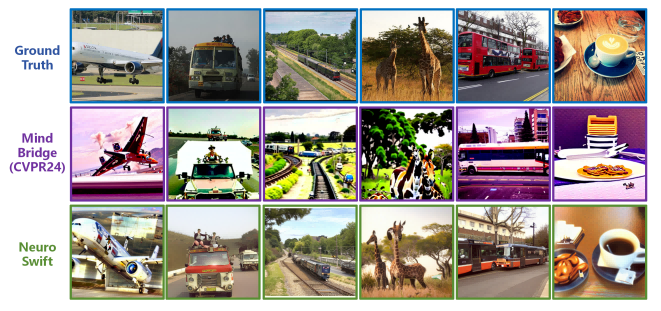}
  \caption{Comparison of our framework with MindBridge\cite{13-5} using 40h of training data in \textbf{single‑subject} mode.}
  \label{fig:5}
\end{figure}

\textbf{1-hour Data.} Figure~\ref{fig:4} shows that NeuroSwift outperforms both MindEye2 and MindTuner in cross‑subject generalization on complex scenes. Take idx2, for example: the scene features a bicycle on a street. MindEye2 misses the bicycle entirely, and although MindTuner brings back the bicycle, it ignores the surrounding traffic. NeuroSwift, however, correctly reconstructs both the bicycle and the traffic flow. In idx3, which depicts a boat on a shore in the lower‑right corner, MindEye2 mistakenly reconstructs a truck, and MindTuner places the boat out on the water. Only NeuroSwift accurately reproduces the boat sitting on the shore. These comparisons highlight NeuroSwift’s superior ability to restore spatial layout and semantic detail at the same time. However, occasional failures do occur. For instance, in idx4 (the “elephant” stimulus), neither MindEye2 nor NeuroSwift manages to recover the elephant.

\textbf{40-hour Data.} When reconstructing using 40 hours of data from a single subject, NeuroSwift outperforms MindBridge. In the six examples shown in Figure~\ref{fig:5}, both methods capture the core semantic content accurately, but MindBridge's reconstructions exhibit spatial deviations from the original stimuli, whereas NeuroSwift achieves precise spatial recovery.

\subsection{Quantitative Evaluation}
To objectively measure reconstruction performance, we conduct quantitative evaluations benchmarking against four state-of-the-art approaches: MindEye2\cite{mindeye2}, MindBridge\cite{13-5}, MindTuner\cite{mindtuner} and BrainGuard\cite{BrainGuard}. We assess low‑level spatial coherence with SSIM \cite{pixcor-ssim} and pixel-wise accuracy via PixCorr \cite{pixcor-ssim}, mid‑level texture consistency using AlexNet(2/5) feature similarity \cite{imagenet-alex}, and high‑level semantics with Incep \cite{inception}, CLIP \cite{uni23-clip}, EffNet-B \cite{10-5-effenet}, and SwAV \cite{swav}. Higher scores are better for SSIM, PixCorr, Alex(2), Alex(5), Incep and CLIP, while lower distances indicate superior performance for EffNet-B and SwAV.
\begin{table}[h]
  \centering
  \includegraphics[width=\linewidth]{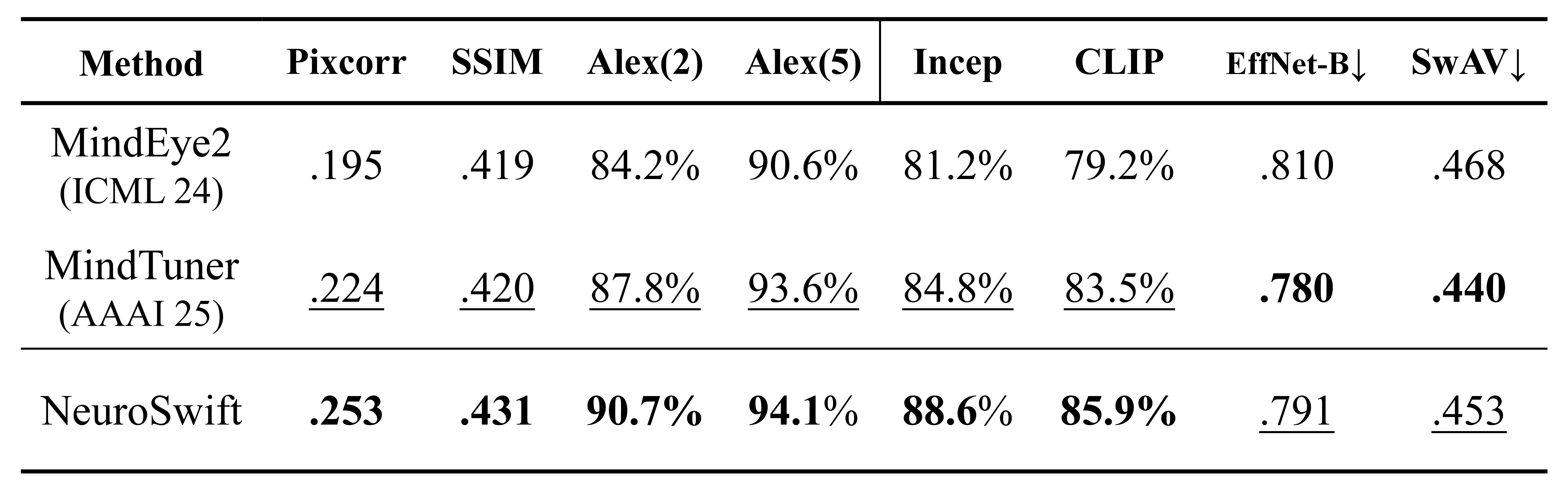}
  \caption{Quantitative comparison using 1h of training data under \textbf{cross‑subject} mode. Bold indicates the best performance, while underline denotes the second-best.}
  \label{tab2}
\end{table}

\textbf{1-hour Data.} As Table~\ref{tab2} shows, with 1-hour of training data using only 3×RTX4090 GPUs under cross‑subject setting, NeuroSwift demonstrates significant advantages. Its PixCorr (0.253) surpasses MindTuner (0.224) and MindEye2 (0.195) by 13\% and 30\% respectively, while SSIM (0.431) also leads competing methods. These results confirm its superior reconstruction of complex scenes, matching our qualitative observations of improved spatial structure recovery. For semantic understanding, NeuroSwift achieves highest in CLIP score (85.9\%) and AlexNet(2) (90.7\%), reflecting its exceptional semantic extraction capabilities. 

The above results indicate that, despite its intermediate parameter count between MindEye2 (64.4M) and MindTuner (76.7M), NeuroSwift (66.3M)  achieves more efficient cross‑subject generalization via its dual‑path architecture (AutoKL+CLIP) combined with subject‑customized brain masks. This design avoids errors caused by imprecise registration inherent in standardized brain templates and enhances downstream processing precision through individualized ROI mapping to deliver cross‑subject performance improvements.
\begin{table}[h]
  \centering
  \includegraphics[width=\linewidth]{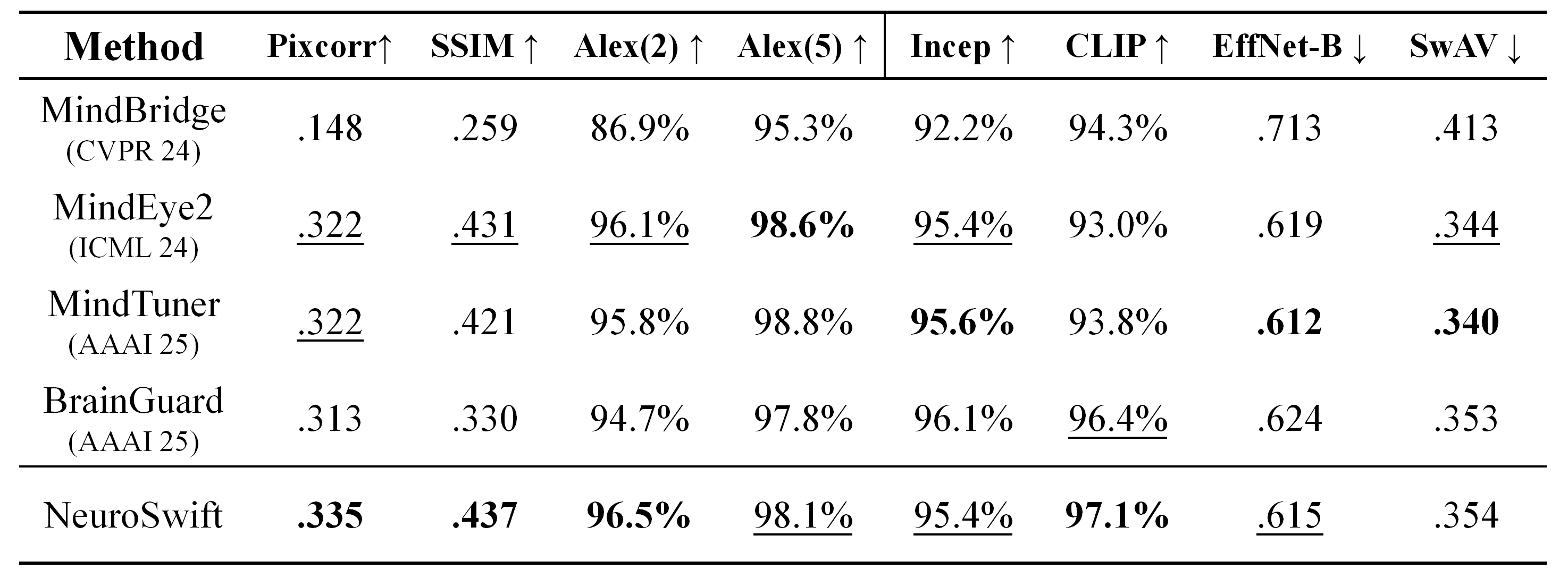}
  \caption{Quantitative comparison using 40h of training data under \textbf{single‑subject} mode.}
  \label{tab3}
\end{table}

\textbf{40-hour Data.} With 40 hours of training data in the single-subject mode, NeuroSwift also demonstrates performance improvements. As shown in Table~\ref{tab3}, it achieves top-performing results in both PixCorr (0.335) and SSIM (0.437), supporting its superior spatial reconstruction capabilities observed in qualitative results. The remarkable breakthrough in CLIP score (97.1\%) directly reflects its precise semantic content recovery (e.g., accurate reconstruction of bicycles and boats). Although EffNet-B (0.615) and SwAV (0.354) slightly underperform the optimal values.

\subsection{Ablation studies}
In ablation studies, we analyze NeuroSwift component contributions through modular removal in four configurations: ``only $\mathit{z}_{\scriptstyle{\textit{pred}}}$'' retains solely the structural pipeline's latent variable for Versatile Diffusion reconstruction without CLIP guidance; ``w/o $\mathit{e}_{\scriptstyle{\textit{txt\_pred}}}$'' removes text embedding guidance, reducing conditioning to image modality; ``w/o $\mathit{e}_{\scriptstyle{\textit{img\_pred}}}$'' disables image embedding guidance, using only text embeddings; ``w/o $\mathit{z}_{\scriptstyle{\textit{pred}}}$'' disables the structural latent variable, initializing diffusion with $ \mathit{z}_{\tau} = \epsilon \sim \mathcal{N}(0, \mathbf{I}) $ and relying solely on text-image embeddings. 
\begin{figure}[h]
  \centering
  \includegraphics[width=0.45\textwidth]{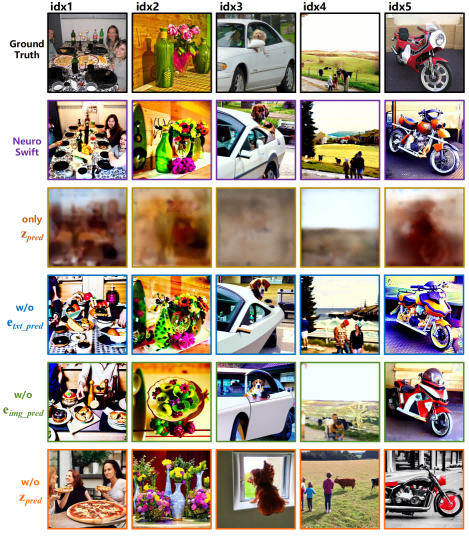}
  \caption{Reconstruction examples from Subj01 with various ablations of the full model.}
  \label{fig:6}
\end{figure}

\textbf{Qualitative results.} As shown in Figure~\ref{fig:6}, the ``only $\mathit{z}{\scriptstyle{\textit{pred}}}$'' configuration preserves low-level spatial features (e.g., contours, layouts) but fails to generate identifiable semantic content due to the absence of CLIP embeddings. Conversely, the ``w/o $\mathit{z}{\scriptstyle{\textit{pred}}}$'' configuration captures primary semantic concepts but exhibits systematic deviations in spatial layouts without the structural prior constraint.

The ``w/o $\mathit{e}_{\scriptstyle{\textit{txt\_pred}}}$'' configuration maintains structural fidelity close to the full model through $\mathit{z}{\scriptstyle{\textit{pred}}}$ and $\mathit{e}_{\scriptstyle{\textit{img\_pred}}}$ synergy, but suffers semantic deterioration (e.g., scene/color inaccuracies), highlighting CLIP text embeddings' role for abstract concepts. Meanwhile, the ``w/o $\mathit{e}_{\scriptstyle{\textit{img\_pred}}}$'' configuration accurately reconstructs semantics but shows reduced structural fidelity in some instances (e.g., object shapes), though $\mathit{z}_{\scriptstyle{\textit{pred}}}$ maintains topological rationality.

Above results reveal that the structural prior ensures the fidelity of low-level spatial topology, CLIP text embeddings constrain abstract semantics, and CLIP image embeddings supplement fine-grained visual features. 

\textbf{Quantitative evaluation.} 
As shown in Table~\ref{tab4}, quantitative experiments demonstrate that when only the structural prior $\mathit{z}{\scriptstyle{\textit{pred}}}$ is retained (w/o $\mathit{e}_{\scriptstyle{\textit{input}}}$), spatial fidelity (SSIM=0.445) and texture consistency (AlexNet(2)=96.8\%, AlexNet(5)=98.5\%) are maximized, but all semantic metrics collapse (Inception=51.2\%, CLIP=54.3\%), confirming the qualitative observation of ``clear contours but blurred semantics." Conversely, removing $\mathit{z}{\scriptstyle{\textit{pred}}}$ (w/o $\mathit{z}{\scriptstyle{\textit{pred}}}$) causes spatial structure to deteriorate dramatically (SSIM falls to 0.237), yet under CLIP's multimodal guidance the semantic scores leap to the top (Inception=96.1\%, CLIP=98.5\%), corresponding to the qualitative case of “semantically accurate but misaligned layout.” 
\begin{table} % H 选项强制表格出现在当前位置
  \centering
  \includegraphics[width=\linewidth]{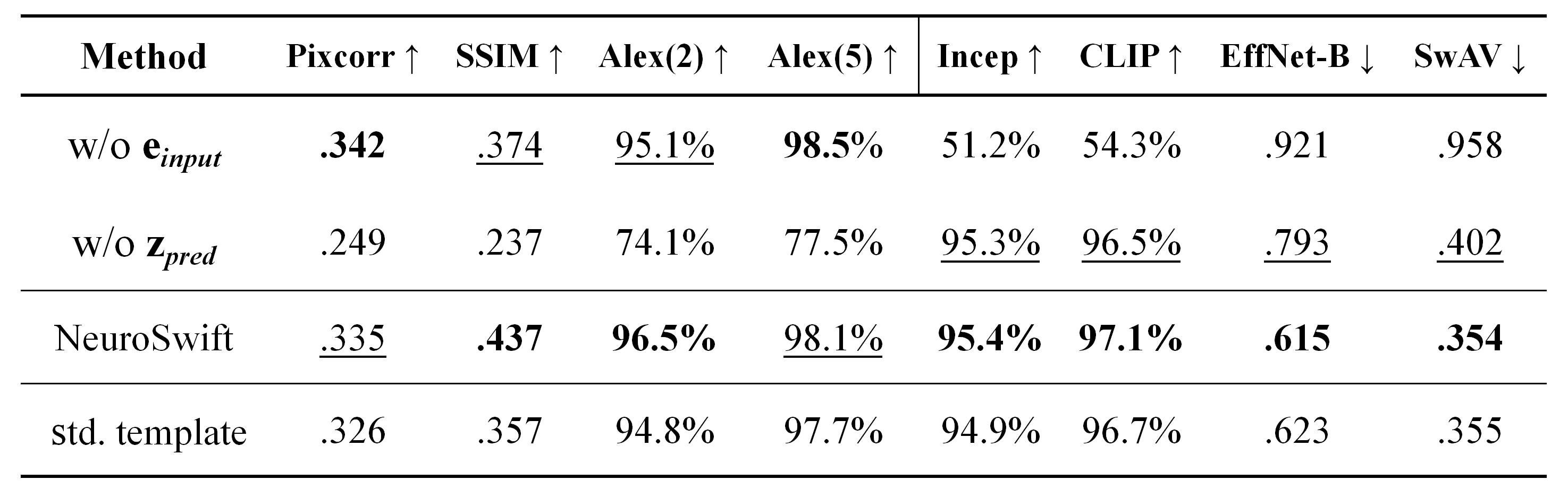}
  \caption{Quantitative comparison of the full model's reconstruction results on Subj01 with its ablated configurations.}
  \label{tab4}
\end{table}

Furthermore, we analyzed the impact of neuroanatomical specificity by replacing manually customized mask with standardized brain template (std. template). While semantic metrics remained competitive, spatial and texture fidelity deteriorated: SSIM plunged by 18.3\% (0.357 vs. 0.437). This gap suggests that using conventional templates may induce boundary ambiguity during spatial normalization, thereby impairing low-level features (SSIM/AlexNet) while retaining only coarse semantic representations.

\subsection{Interpretability of NeuroSwift}
\begin{figure}[h]
  \centering
  \includegraphics[width=0.5\textwidth]{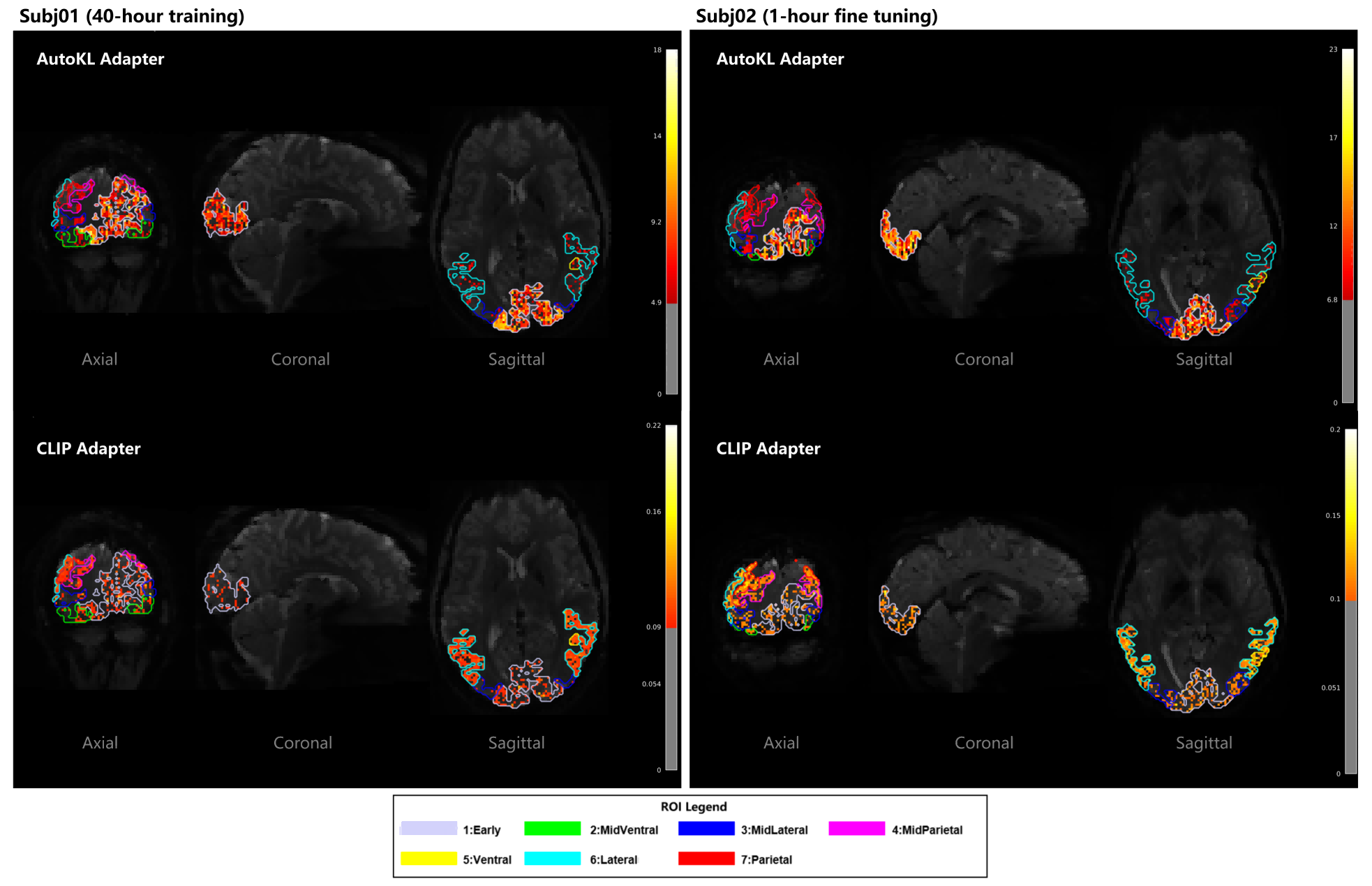}
  \caption{Spatial distribution of brain‑region contributions in Subj01 for the AutoKL Adapter (a) and CLIP Adapter (b).}
  \label{fig:7}
\end{figure}

To investigate the importance of different brain regions in the NeuroSwift for visual reconstruction, we extracted the weight matrices of the AutoKL and CLIP Adapters’ fully connected layers and computed each voxel's L2 norm as its activation contribution. Figure~\ref{fig:7} shows the results obtained by training on 40 hours of data from Subj01 and fine-tuning with 1 hour of data from Subj02. The cross-subject fine-tuning results for Subj05 and Subj07 are provided in Appendix~\ref{apen.weights}.

The AutoKL Adapter weight map shows prominent activation clusters in ``Early" regions (V1-V3), indicating higher contributions to structural generation, consistent with its role in preserving low-level features like edges and textures. Conversely, it displays sparse activation in higher-order cortices (``Lateral"/``Parietal"). In contrast, the CLIP Adapter weight map exhibits continuous ``lava flow" activation across ``Ventral", ``Lateral", and ``Parietal" regions, paralleling its capacity to capture semantic information through multimodal integration. Conversely, sparse activations in primary/intermediate regions (Early, Midventral, etc.) indicate their limited semantic encoding contributions.

\section{Conclusion and Limitations}
NeuroSwift's dual-pathway architecture overcomes complex scene limitations: an AutoKL Adapter preserves low-level structural details while a biologically inspired CLIP Adapter extracts core semantics via synthetic images and text captions. Using manually delineated ROIs and pretraining with lightweight fine-tuning for new subjects, it achieves state-of-the-art cross-subject generalization with just one hour of data per subject on three RTX4090 GPUs.

Current limitations include dataset-specific validation, as significant differences in experimental design and signal acquisition restrict generalizability to the NSD dataset\cite{8-14}. Future work will conduct cross-dataset validation on BOLD5000\cite{bold5000} to enhance robustness.

\begin{acks}
This work was supported by the Strategic Priority Research Program of Chinese Academy of Sciences (XDB0930302), and the Guangdong Provincial Key Laboratory of Multimodality Non-Invasive Brain-Computer Interfaces (2024B1212010010).
\end{acks}


%\bibliographystyle{ACM-Reference-Format}

\bibliographystyle{ACM-Reference-Format}
\bibliography{sample-base}

\end{document}